# Probabilistic Data Quality Assessment for Structural Monitoring Data Via Outlier-resistant Conditional Diffusion Model

Qi Li[1,2], Yong Huang[1,2*], Hui Li[1,2]

[1.] Key Lab of Smart Prevention and Mitigation of Civil Engineering Disasters of the Ministry of Industry and Information Technology, School of Civil Engineering, Harbin Institute of Technology, Harbin, 150090, China

[2.] Key Lab of Structures Dynamic Behavior and Control of the Ministry of Education, Harbin Institute of Technology, Harbin, 150090, China

## Abstract

Data quality assessment is an essential step that ensures the reliability of the subsequent structural health monitoring (SHM) tasks. This study proposes a prediction deviation-based SHM data quality assessment method using a univariate implicit autoregressive model, enabling outlier diagnosis and data cleaning. The proposed conditional diffusion model (CDM) augments the standard diffusion model with a conditional embedding module to incorporate temporal context, quartile normalization to mitigate distribution skew, and a Huber loss to enhance robustness against outliers. Within this univariate implicit autoregressive framework, each data point is assigned an outlier probability, quantifying its degree of "outlier-ness", and a global quality evaluation score is computed to characterize the overall dataset quality. Extensive case studies utilizing operational data from real-world structures demonstrate that the proposed framework significantly improves the accuracy of data quality assessment, outperforming other strong baselines representative of clustering, isolation-based, and deep reconstruction methods. The effectiveness and robustness of the proposed framework are further demonstrated by the findings of ablation experiments and hyperparameter analysis.

# 1.Introduction

As global transport networks expand, long-lived assets such as bridges and high-speed railways undergo cumulative environmental and traffic-induced deterioration with potentially catastrophic consequences, making effective structural health monitoring (SHM) (Bao et al., 2019; Bao & Li, 2021; Gharehbaghi et al., 2022; Xu et al.,2024) indispensable. SHM deploys environmental sensors (e.g., temperature, wind speed, wind direction) and structural response sensors (e.g., acceleration, strain, relative displacement) on the asset in order to collect operational data. Subsequent analyses utilize these data to identify potential damage, fatigue, and other safety risks, thereby helping to extend service life. The accuracy and reliability of the data are of paramount importance, because clean and reliable datasets are prerequisites for the successful implementation of subsequent tasks such as structural condition assessment (Ni et al., 2020; Wang et al., 2024; Wei et al., 2025), damage detection (Wan et al., 2023; Xue et al., 2023; Shi et al., 2025) and early warning (Wang et al., 2022; Liu et al., 2022). In practice, outliers are unavoidable due to environmental impacts, sensor faults, and unstable data transmission (Bao et al., 2019). Consequently, data quality assessment, also termed data outlier detection by some researchers (Aslan & Onut, 2022; Hussain & Zhang, 2025; Papastefanopoulos et al., 2025; Yuen & Ortiz, 2017), has become a central task in current SHM research.

Numerous studies on data quality assessment have been conducted. Several data quality assessment methods leverage cluster analysis, classifying data points in low-density regions as outliers. For example, Aslan & Onut (2022) employed the Density-Based Spatial Clustering of Applications with Noise (DBSCAN) algorithm to identify outliers and extreme events in ground-based PM10 measurements. The algorithm requires two parameters, a neighborhood radius and the minimum number of points within that radius to form a cluster. It then classifies points as core points, boundary points, or noise based on density connectivity, and it performs well in practice. Shrifan et al. (2022) proposed an improved K-means algorithm that incorporate Tukey's rule together with a novel distance metric. The method determines whether the data distribution is left-skewed, right-skewed, or symmetric about the mean, which enables adaptive removal of outliers. Although effective, cluster-based data quality assessment methods require predefined hyperparameters (e.g., neighborhood radius, minimum cluster size, and cluster count), which makes performance highly sensitive to the chosen values.

Isolation mechanism methods, such as Isolation Forest and Extended Isolated

Forest, which replace explicit modeling of normal distributions with "random partition - fast isolation", have become one of the mainstream paradigms for unsupervised anomaly detection. Liu et al. (2008) first proposed the Isolation Forest method, which measures the anomaly degree by randomly dividing the space and isolating the path length. It has the advantages of approximately linear complexity and low memory requirements. Fang et al. (2023) conducted unsupervised anomaly detection on sensor monitoring data for deep foundation pit construction using EIF, and combined with Variational Mode Decomposition to achieve adaptive denoising. Verification on real-life data shows that this method can effectively improve data quality and is helpful in supporting subsequent geotechnical safety assessment and decision-making. Despite their scalability and simplicity, isolation-based methods are sensitive to threshold choices and training-set contamination, and depend heavily on feature scaling or representation.

Owing to its powerful representation learning and capacity to model highly non-linear decision boundaries, deep learning methods have also been explored for data quality assessment in recent years. Xiao et al. (2024) proposed an autoencoder-based method for assessing data quality in bridge health monitoring that adaptively determines evaluation thresholds by calculating Euclidean distance between compressed latent space representations. Deng et al. (2023) leveraged a continuous wavelet transform to extract time-frequency features and subsequently constructed a convolutional neural network to assess data quality, and they demonstrated cross-object applications. Pan et al. (2023) developed a transfer learning-based anomaly detection method that exploits similar in anomaly patterns across different bridges, enabling knowledge transfer for high-precision anomaly detection in bridge groups. However, the output of these approaches are typically continuous time series data and do not provide precision at the level of individual data point. They are less suitable when sampling frequency is low and monitoring data are sparse. Furthermore, they still rely on manually labeled datasets for supervised methods.

Data quality assessment methods based on model prediction deviations constitute another significant category, encompassing both machine learning and statistical models. These methods identify data points exhibiting large prediction deviations as outliers. Hussain & Zhang (2025) proposed a machine learning-based anomaly detection method for pipeline internal inspection data. Their approach employs an improved interquartile range technique, although optimal performance still requires user-defined thresholds. Yuen & Ortiz (2017) proposed a robust regression and outlier detection method for correlated data. It explicitly accounts for interdependence using the minimum volume ellipsoid method to quantify the outlier probability for each data point, and it was validated on both simulated and real-world data. Li et al. (2024) proposed a Bayesian extreme learning machine–based method for outlier detection in long-term high-speed rail track monitoring data. It computes outlier probabilities from posterior mean uncertainty and noise, and substantially improves prediction accuracy after removing detected outliers. Nevertheless, these approaches typically rely on regression relationships among different variables. If either the input or the output contains an outlier, the entire input-output data pair is flagged as anomalous. In practical SHM systems, the profusion and coupling of variables make it difficult to define input–output mappings for such cross-correlation–based regression. Owing to persistent environmental forcing and the modal decay of structural responses, data collected by SHM systems typically exhibit strong serial dependence (Farrar & Worden, 2012; Yao & Pakzad, 2012). It is therefore often more natural to model autoregression behavior within individual time series rather than perform cross-regression among disparate variables (Qu et al., 2022).

Beyond the difficulty of specifying input-output mappings, prediction deviation methods suffer from another limitation, namely that they require highly accurate predictions to enable precise data quality assessment. Consequently, they typically rely on pre-cleaning the training data or further screening a relatively reliable subset for model training (Yuen & Ortiz, 2017; Qu et al., 2022; Li et al., 2024), which further limits their practicality. With recent advances in generative artificial, diffusion models (Ho et al., 2020) have become a promising avenue for this problem. Diffusion models are intrinsically well suited to SHM data quality assessment because their forward–reverse noising paradigm learns a noise-aware, multi-scale representation of the data distribution with calibrated uncertainty and thus provides inherent robustness to sensor noise, impulsive spikes, and heavy-tailed perturbations. Beyond image generation, diffusion models have been applied to time-series forecasting (Rasul et al., 2021), physics-informed prediction (Jadhav et al., 2023), engineering design (Gu et al., 2024), and dynamic-response reconstruction (Shu et al, 2025). Their progressive denoising mechanism enables stable training without mode collapse, provides explicit control of noise levels that mirrors real-world corruption, and yields full probabilistic distribution through the learned reverse process. Nevertheless, the robustness of diffusion model to low-quality measurements and their use as a probabilistic engine for SHM data quality assessment remain largely underexplored.

To fill these gaps, we propose a probabilistic data quality assessment method for SHM data via outlier-resistant conditional diffusion model (CDM). Specifically, we develop a CDM for autoregressive time-series forecasting that integrates conditional embeddings to encode contextual and exogenous information, quartile normalization to mitigate heavy-tailed contamination, and a Huber loss to reduce sensitivity to outliers during training. Since the conditional embedding module leverages past contextual information to guide the generation process, it can also be considered an implicit autoregressive method. Within this probabilistic framework, we compute pointwise outlier probabilities by jointly accounting for various sources of uncertainties, yielding a quantitative measure of data quality at each datapoint; these probabilities are further aggregated into a global quality evaluation score. The proposed method is validated using real-world data from a bridge and a high-speed railway track monitoring system.

The remainder of the paper is organized as follows. In Section 2, we introduce the outlier-resistant conditional diffusion model and the associated data quality assessment method. Section 3 presents two illustrative applications to demonstrate the effectiveness and practicality. We further discuss practical deployment challenges and mitigation strategies. Section 4 summarizes the main contributions and directions for future work.

## 2. The proposed approach

### 2.1 Overview

Let $D = \{x_1^0, x_2^0, \ldots, x_N^0, x_{N+1}^0, \ldots, x_{N+L}^0\}$ denote the univariate structural health monitoring time series data, where $x_N^0$ is the monitoring data corresponding to the $N$-th time moment. The objective of data quality assessment is to reliably discriminate normal from anomalous observations, thereby preventing outliers from biasing subsequent SHM analysis and decision-making. To accomplish this goal, we propose a novel probabilistic method for detecting outliers with the associated discrimination criterion in the form of outlier probability. This approach can estimate both the probability that each data point is an outlier (i.e., local data-quality assessment) and, from these probabilities, a quality score for the entire time series (i.e., global quality assessment). The framework of the proposed method is schematically illustrated in Fig 1. It comprises two main components: an outlier-resistant conditional diffusion model for probabilistic prediction, and an outlier probability compute and data quality assessment module. These components are detailed in subsequent subsections.

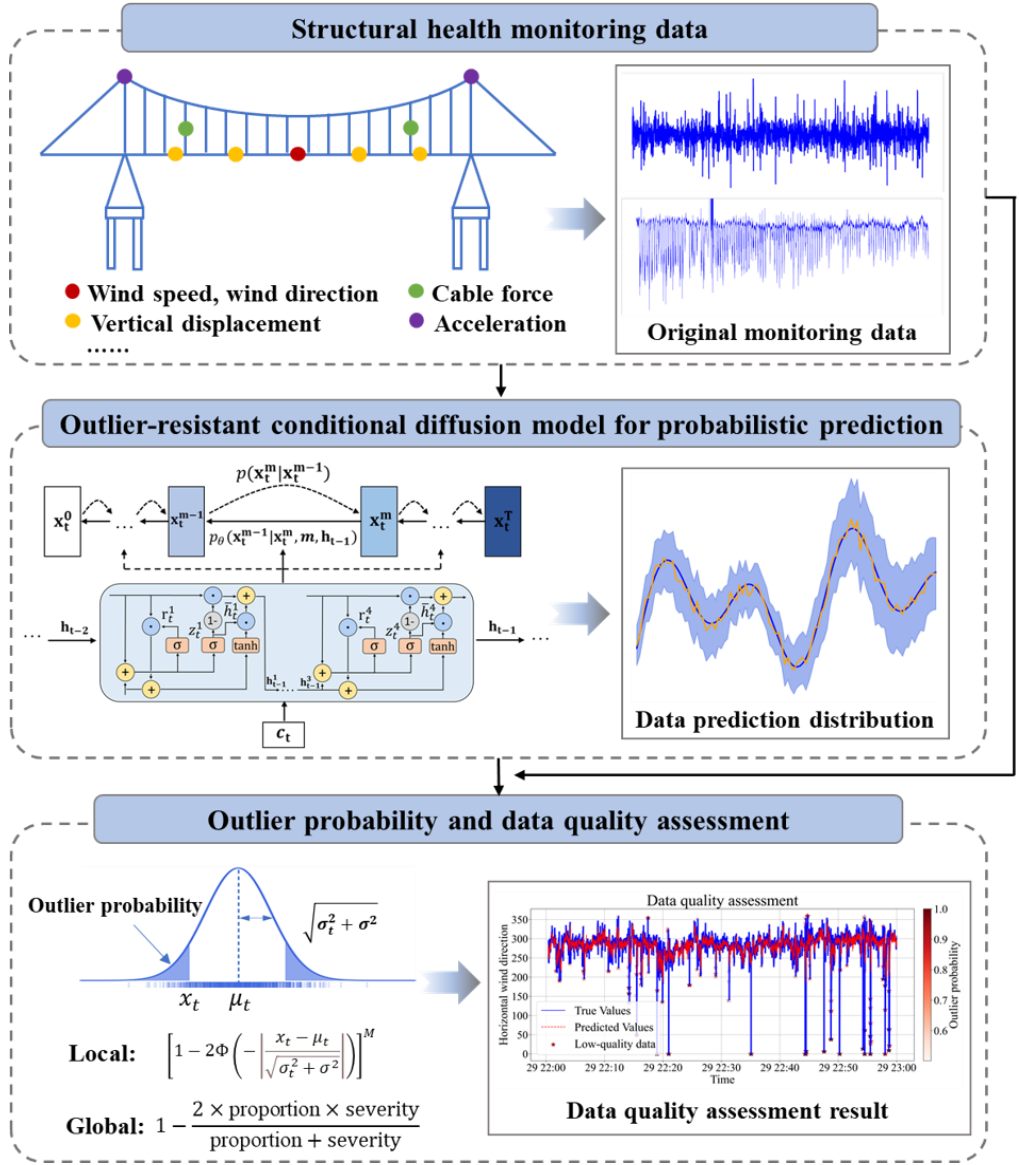

**Fig. 1** Flowchart of the proposed method for data quality assessment

## 2.2 Outlier-resistant conditional diffusion model for probabilistic prediction

The Denoising Diffusion Probabilistic Model (DDPM) (Ho et al., 2020) is a generative framework that draws inspiration from nonequilibrium thermodynamics. It generates data by gradually corrupting samples with noise and then learning a reverse diffusion (denoising) process. Its core is the reverse diffusion process, which reconstructs the underlying structure of the data from noise.

In the forward diffusion process, the original monitoring data $x_t^0$ is gradually corrupted with Gaussian noise until the data eventually becomes pure noise. Here, $x_t^0$ denotes the monitoring data at time step $t$. The amount of injected Gaussian noise at step $m$ is controlled by the noise coefficient $\beta_m$, and the process is modeled as a Markov chain:

$$x_t^m = \sqrt{1-\beta_m^2}\, x_t^{m-1} + \beta_m \varepsilon \tag{1}$$

$$x_t^m = \bar{\alpha}_m x_t^0 + \bar{\beta}_m \varepsilon \tag{2}$$

where $x_t^m$ is the data after $m$ diffusion steps, $\bar{\alpha}_m = \sqrt{1-\beta_m^2} \cdot \ldots \cdot \sqrt{1-\beta_1^2}$, $\bar{\beta}_m = \sqrt{1-\bar{\alpha}_m^2}$, and $\varepsilon \sim \mathcal{N}(0, \mathrm{I})$. The sequence $\{\beta_m\}_{m=1}^T$ is chosen to be monotonically increasing over the diffusion steps. Early steps inject less noise, later steps inject more, until the sample becomes pure noise. The linear schedule for $\beta_m$ is as follows:

$$\beta_m = \beta_{min} + (\beta_{max} - \beta_{min}) \cdot \frac{m-1}{T-1} \tag{3}$$

where $\beta_{min}$ and $\beta_{max}$ are the preset minimum and maximum values, respectively, and $T$ is the total number of diffusion steps.

The reverse generation process starts from $x_t^T \sim \mathcal{N}(0, \mathrm{I})$ and iteratively removes noise to recover $x_t^0$. The process is principally concerned with the derivation of the diffusion state at step $m-1$, $x_t^{m-1}$, from the diffusion state at step $m$, $x_t^m$. In accordance with the forward diffusion process delineated above, the likelihood function for the diffusion state at step $m$, given the diffusion state at step $m-1$ is as follows:

$$p(x_t^m | x_t^{m-1}) = \mathcal{N}(x_t^m; \sqrt{1-\beta_m^2}\, x_t^{m-1}, \beta_m^2 \mathrm{I}) \tag{4}$$

As demonstrated in Eq. (2), given $x_t^0$, the prior probability density function (PDF) of the diffusion state at step $m-1$, $x_t^{m-1}$, is Gaussian:

$$p(x_t^{m-1}|x_t^0) = \mathcal{N}(x_t^{m-1}; \bar{\alpha}_{m-1}x_t^0, \bar{\beta}_{m-1}^2 I) \tag{5}$$

Utilizing the likelihood function in Eq. (4) and the prior PDF in Eq. (5), the posterior PDF can be derived by Bayes' Theorem (Huang et al., 2019; Meng et al., 2025):

$$p(x_t^{m-1}|x_t^m, x_t^0) = \frac{p(x_t^m|x_t^{m-1})p(x_t^{m-1}|x_t^0)}{p(x_t^m|x_t^0)} = \mathcal{N}(x_t^{m-1}; \mu_m(x_t^m, x_t^0), \frac{1}{\tilde{\beta}_m^2} I) \tag{6}$$

where the posterior mean and precision parameter are:

$$\mu_m(x_t^m, x_t^0) = \frac{\bar{\alpha}_{m-1}\beta_m^2}{\bar{\beta}_m^2} x_t^0 + \frac{\alpha_m \bar{\beta}_{m-1}^2}{\bar{\beta}_m^2} x_t^m \tag{7}$$

$$\tilde{\beta}_m^2 = \frac{\bar{\beta}_m^2}{\bar{\beta}_{m-1}^2 \beta_m^2} \tag{8}$$

The normalization constant for the posterior probability density function, known as the evidence function, is:

$$p(x_t^m|x_t^0) = \mathcal{N}(x_t^m; \bar{\alpha}_m x_t^0, \bar{\beta}_m^2 I) \tag{9}$$

It is imperative to acknowledge that Eq. (7), Eq. (8) and Eq. (9) have been derived under the supposition that $x_t^0$ is given. Nevertheless, $x_t^0$ is the final result that should be generated and is not available during the reverse generation process. In order to remove this dependency on $x_t^0$, an alternative strategy involves employing a neural network to approximate the relationship between $x_t^m$ and $x_t^0$. In the event that $x_t^0$ can be predicted from $x_t^m$, it follows that the reverse generation process can rely solely on $x_t^m$:

$$p(x_t^{m-1}|x_t^m) \approx p_\theta(x_t^{m-1}|x_t^m, x_t^0 = \bar{\mu}(x_t^m)) \tag{10}$$

From Eq. (2), it can be deduced that the following equation governs the relationship between $x_t^0$ and $x_t^m$:

$$x_t^0 = \frac{1}{\bar{\alpha}_m}(x_t^m - \bar{\beta}_m \boldsymbol{\varepsilon}) \tag{11}$$

Based on the form of $x_t^0$, the prediction result $\bar{\mu}(x_t^m)$ of the neural network is chosen as:

$$\bar{\mu}(x_t^m) = \frac{1}{\bar{\alpha}_m}(x_t^m - \bar{\beta}_m \boldsymbol{\varepsilon_\theta}(x_t^m, m)) \tag{12}$$

The loss function is conventionally designated as the $\ell_2 - $ norm, i.e., $\|x_t^0 - \bar{\mu}(x_t^m)\|_2^2$, where $x_t^0$ represents the original data and $x_t^m$ represents the noisy data. Therefore, the network essentially operates as a denoising network.

In order to enhance the robustness of the model against outliers and ensure its applicability to SHM data, three improved strategies have been adopted, based on the traditional diffusion model. Specifically, we add a conditional embedding network, apply quartile normalization and train with the Huber loss function. The conditional embedding network module employs a Recurrent Neural Network, specifically utilizing Gated Recurrent Units (Dey & Salem, 2017), which offer faster training speeds and fewer parameters compared to Long Short-Term Memory cells. This module enables the conditioning of time series data generation not only on the temporal characteristics of the current time step but also on the data characteristics within the context window, thereby implicitly incorporating autoregressive dependencies. A schematic of the conditional diffusion model is illustrated in Fig. 2.

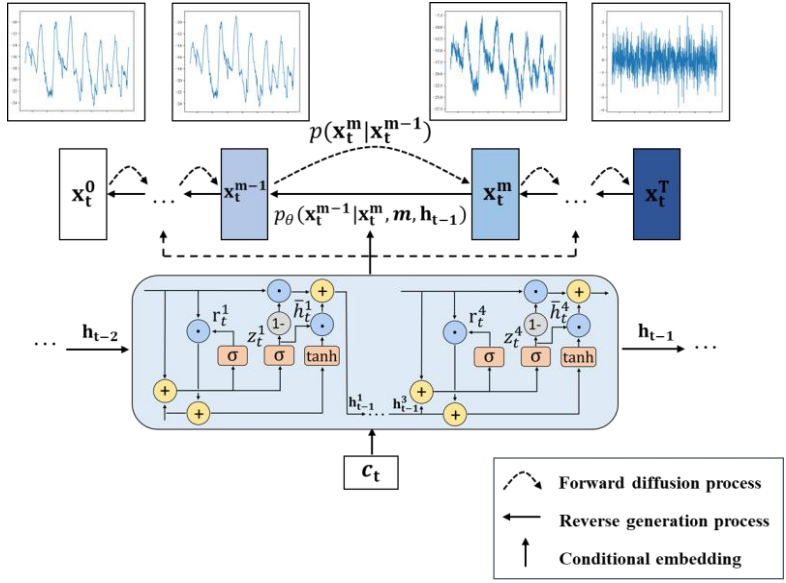

**Fig. 2** The structure of the conditional diffusion model

The network consists of four hidden layers, with each hidden layer comprising 30 nodes. The covariate $c_t$ is composed of time-dependent features at time step $t$ (such as the date, the day of the week, hour, and minute), as well as lag features from the context window. The covariate $c_t$, along with the hidden state $h_{t-1} = \{h_{t-1}^1, h_{t-1}^2, h_{t-1}^3, h_{t-1}^4\}$ from the previous time step $t-1$, serves as the input to the conditional network. The output of the first layer is the hidden state $h_t^1$, which is computed through the following equations:

$$r_{t-1}^1 = \sigma(W_r^1 \cdot [h_{t-1}^1; c_t] + b_r^1) \tag{13}$$

$$z_{t-1}^1 = \sigma(W_z^1 \cdot [h_{t-1}^1; c_t] + b_z^1) \tag{14}$$

$$\tilde{h}_{t-1}^1 = tanh(W_h^1 \cdot (r_{t-1}^1 \odot [h_{t-1}^1; c_t]) + b_h^1) \tag{15}$$

$$h_t^1 = (1 - z_{t-1}^1) \odot h_{t-1}^1 + z_{t-1}^1 \odot \tilde{h}_{t-1}^1 \tag{16}$$

The function $\sigma(x) = \frac{1}{1+e^{-x}}$ represents the sigmoid activation function, and $tanh(x) = \frac{e^x - e^{-x}}{e^x + e^{-x}}$ represents the Tanh activation function. The symbol $\odot$ denotes element-wise multiplication. The computation type for the subsequent layers is similar, with the only difference being that the input to each layer is the output of the previous layer. The set $\mathbf{R_\theta} = \{\mathbf{W_r^i}, \mathbf{W_z^i}, \mathbf{W_h^i}, \mathbf{b_r^i}, \mathbf{b_z^i}, \mathbf{b_h^i}\}_{i=1}^{4}$ refers to the model parameters of the conditional embedding network, which are initialized randomly from a standard normal distribution and optimized during the training process. The initial hidden state is given by $\mathbf{h_0} = \mathbf{0}$.

The denoising network module takes the hidden state as an additional input and comprises eight residual blocks with skip connections. The network architecture is illustrated in Fig. 3. At each iteration, a diffusion step $m$ is chosen from the set $\{1, 2, \dots, T\}$, and the network model parameters are optimized by gradient descent. The learnable parameters consist of two sets: $\mathbf{R_\theta}$ for the conditional embedding network and $\mathbf{\varepsilon_\theta}$ for the denoising network.

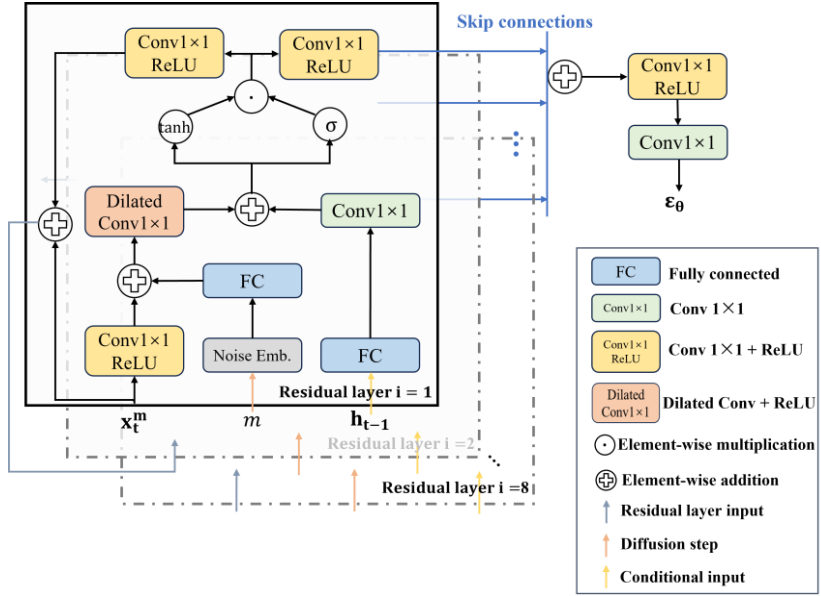

**Fig. 3** The structure of the denoising network

The second improvement strategy concerns the choice of data normalization method. This study utilizes quantile normalization instead of the prevalent min-max Normalization or Z-score normalization methods. The scaled value is computed as:

$$x_{scaled} = \frac{x_t - x_{50\%}}{x_{75\%} - x_{25\%}} \tag{17}$$

where $x_{25\%}$, $x_{50\%}$ and $x_{75\%}$ denote the 25$^{th}$, 50$^{th}$ and 75$^{th}$ percentiles computed on the training set, respectively. This method centers each feature by its median and scales by its interquartile range, without forcing identical feature distributions across samples. It effectively mitigates the impact of extreme outliers, thereby enhancing model robustness and accuracy. Finally, we optimize with the Huber loss instead of the standard $\ell_2$-norm:

$$Loss_{huber} = \sum_{m=1}^{T} \begin{cases} (\boldsymbol{\varepsilon} - \boldsymbol{\varepsilon_\theta}(\sqrt{\bar{\alpha}_m}x_i^0 + \sqrt{1-\bar{\alpha}_m}\boldsymbol{\varepsilon}, \mathbf{h_{t-1}}, m))^2, & |\boldsymbol{\varepsilon} - \boldsymbol{\varepsilon_\theta}(\sqrt{\bar{\alpha}_m}x_i^0 + \sqrt{1-\bar{\alpha}_m}\boldsymbol{\varepsilon}, \mathbf{h_{t-1}}, m)| \leq 1 \\ |\boldsymbol{\varepsilon} - \boldsymbol{\varepsilon_\theta}(\sqrt{\bar{\alpha}_m}x_i^0 + \sqrt{1-\bar{\alpha}_m}\boldsymbol{\varepsilon}, \mathbf{h_{t-1}}, m)|, & \text{otherwise} \end{cases}$$

(18)

The Huber loss is quadratic near zero (smooth gradients) and becomes linear for large residuals, limiting outlier influence.

## 2.3 Outlier probability and data quality assessment

After training, given the hidden state $\mathbf{h_{t-1}}$, $x_t^{m-1}$ can be sampled from the distribution based on Eq. (7), Eq. (8) and Eq. (12):

$$x_t^{m-1} = \frac{1}{\sqrt{1-\beta_m^2}}\left(x_t^m - \frac{\beta_m^2}{\sqrt{1-\bar{\alpha}_m^2}}\boldsymbol{\varepsilon_\theta}(x_t^m, \mathbf{h_{t-1}}, m)\right) + \frac{z}{\tilde{\beta}_m} \tag{19}$$

where $z \sim \mathcal{N}(0, \mathbf{I})$ for $m = 2, 3, \ldots, T$, and $z = 0$ for $m = 1$. After completing the reverse generation process for $T$ steps, the model's predicted output $x_t^{0,i}$ at time step $t$ is obtained under the given condition. By sampling $M$ samples from standard

Gaussian noise, the predicted mean $\mu_t$ and standard deviation $\sigma_t$ at this time step are given by:

$$\mu_t = \frac{1}{M}\sum_{i=1}^{M} x_t^{0,i} \quad (20)$$

$$\sigma_t = \sqrt{\frac{1}{M-1}\sum_{i=1}^{M}(x_t^{0,i} - \mu_t)^2} \quad (21)$$

where $x_t^{0,i}$ denotes the $i$-th generated prediction sample. Since sampling progressively denoises an initial Gaussian noise, we approximate the predictive distribution as a Gaussian distribution $p(x_t) \sim \mathcal{N}(\mu_t, \sigma_t^2)$.

Many existing data quality assessment methods in regression analysis are based on the deviations of the predictions, where the threshold for identifying outliers is often set arbitrarily. For instance, some methods use the criterion $|\varepsilon|/\sigma > 3$ for outlier detection (Alimohammadi & Chen, 2022). It goes without saying that such a threshold has a significant impact on the data quality assessment results. In this paper, we propose an alternative approach through the definition and computation of outlier probability for each data point within the time series. This probability integrates data noise, modelling errors and uncertainty in model prediction, offering a more nuanced, data-driven criterion for data quality assessment.

Taking into account both data noise and modelling errors, the relationship in the time series forecasting model can be expressed as:

$$\mathbf{Y} = \text{CDM}(\mathbf{X} \oplus \mathbf{T}) + \boldsymbol{\varepsilon} \quad (22)$$

where $\oplus$ represents the concatenation of the feature dimension of the matrix along the row direction, $\mathbf{X} \oplus \mathbf{T} = [\mathbf{x}_{C+1} \oplus \mathbf{t}_{C+1}, \ldots, \mathbf{x}_N \oplus \mathbf{t}_N]^T \in \mathbb{R}^{(N-C) \times (C+f)}$ and $\mathbf{x}_t \oplus \mathbf{t}_t =$ $[x_{t-C}, ..., x_{t-1}, t_1, ..., t_f]^T \in \mathbb{R}^{C+f}$ ($N$ denotes the number of time instants, $C$ the context window length, and $f$ the dimension of time-dependent features) are the input data matrix for $(N - C)$ time instants and data vector for the $i$-th time instant, CDM(·) refers to the conditional diffusion model for time series forecasting, and $\mathbf{Y} = [x_{C+1}, ..., x_N]^T \in \mathbb{R}^{(N-C) \times 1}$ is the corresponding output data vector of the model. $\boldsymbol{\varepsilon} = [\varepsilon_{C+1}, ..., \varepsilon_N]^T$ is the $(N - C) \times 1$ prediction-error vector, which can be modeled as an $(N - C)$-variate zero-mean Gaussian distribution $\mathcal{N}(\mathbf{0}, \sigma^2 \mathbf{I}_{N-C})$ inspired by the principle of maximum entropy (Huang et al., 2017) and the $\sigma^2$ is the prediction-error variance parameter. It is assumed that the correlations between the time series have been fully learned during the model's training phase, and thus, the prediction errors $\{\varepsilon_{C+1}, ..., \varepsilon_N\}$ for each data points are modelled as mutually independent.

Due to the presence of outliers, a resampling-based method is employed to robustly estimate the prediction error variance parameter. The variance parameter is estimated by repeatedly sampling random subsets of the data $L$ times from the residual dataset:

$$\sigma^2 = \frac{1}{L}\sum_{i=1}^{L}\left[\frac{1}{S_i - 1}\sum_{r=1}^{S_i}(\varepsilon_r - \mu_{S_i})^2\right] \tag{23}$$

where $\varepsilon_r = x_r - \mu_r$ is the residual of r-th data point, $S_i$ denotes a subset of the residual data which formed by randomly selecting a fraction of the residual data, and $\mu_{S_i}$ represents the mean of this subset. Given the predicted mean, variance, and the variance of the prediction errors, the predictive distribution of the observation at time step $t$ remains a Gaussian distribution due to the additivity property of the Gaussian

distribution:

$$p(x_t) = \mathcal{N}(x_t|\mu_t, \sigma_t^2 + \sigma^2) \tag{24}$$

Outliers are typically defined as data points that exhibit substantial deviations from the expected or predicted behavior (Aggarwal, 2017). In this paper, the degree of "outlier-ness" is defined as the probability that all $M$ data points generated through prediction will be closer to the predicted mean $\mu_t$ than to the actual observation, given the probability model $p(x_t)$ of the monitoring data $x_t$. This probability indicates the likelihood that the absolute residuals of all $M$ data points are smaller than the current $|\varepsilon_t|$. Given the parameters $\mu_t, \sigma_t^2$ and $\sigma^2$, the outlier probability for the $t$-th data point $x_t$ is computed as:

$$P_o(x_t|\mu_t, \sigma_t^2, \sigma^2) = [1 - 2\Phi(-\left|\frac{x_t - \mu_t}{\sqrt{\sigma_t^2 + \sigma^2}}\right|)]^M \tag{25}$$

where $\Phi(\cdot)$ denotes the cumulative distribution function of the standard normal distribution. If $P_o(x_t|\mu_t, \sigma_t^2, \sigma^2) > 0.5$, it indicates that the probability of $x_t$ being an outlier is greater than that of being a normal data point. In this case, $x_t$ should be classified as an outlier and moved to the low-quality dataset **U**; otherwise, it will be retained as a normal data point.

Following the determination of outlier probabilities for individual data points (local quality assessment), the quality evaluation score ($QES$) for the entire time series dataset (global quality assessment) is defined by integrating both the proportion of outliers within the dataset and the severity of their anomalies:

$$QES = 1 - \frac{2 \times \frac{Q}{k(N-C)} \times \frac{1}{Q}[\sum_{i=1}^{Q} P_o(x_i|\mu_i,\sigma_i^2,\sigma^2)]}{\frac{Q}{k(N-C)} + \frac{1}{Q}[\sum_{i=1}^{Q} P_o(x_i|\mu_i,\sigma_i^2,\sigma^2)]} \quad (26)$$

where $Q$ is the number of identified outliers, that is, the number of data points corresponding to $P_o(x_i|\mu_i,\sigma_i^2,\sigma^2) > 0.5$; $k \in (0,1)$ is a dimensionless scaling factor to rescale the outlier-proportion term. The motivation is to make the dataset-level deliberately conservation. In other words, $k$ sets a reference level such that, once the detected outlier proportion becomes large, $QES$ is penalized more strongly. $QES$ is defined analogously to the F1-score, combining the proportion of outlier and their anomaly severity via a harmonic-mean-style aggregation. As either the proportion of outlier or the aggregate anomaly severity increases, $QES$ decreases (lower is worse quality, higher is better). When taking $QES$ as the optimization objective, the application of multiple iterations enables the effective identification of all possible instances of outliers.

## 2.4 Proposed algorithm

The data quality assessment algorithm for structural monitoring data is summarized in Fig. 4, and is as follows:

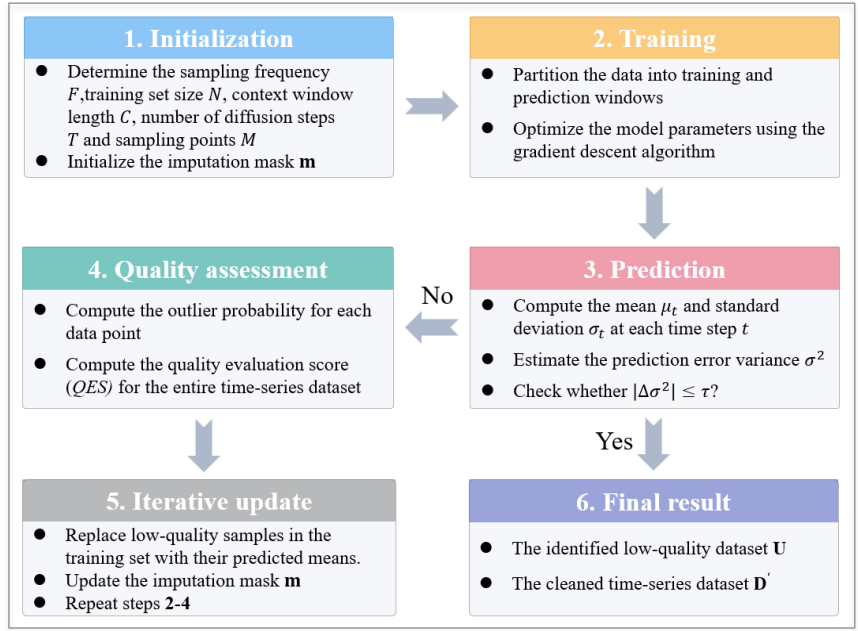

**Fig. 4** Workflow of the time-series cleaning framework with data quality assessment

1. For a univariate time-series dataset **D**: (a) determine the sampling frequency $F$, training set size $N$, context window length $C$, number of diffusion steps $T$ and number of sampling points $M$; (b) initialize the imputation mask $\mathbf{m} = [0,\dots,0]^T$.

2. The $N$ data points constituting the training set data are segmented into $N - C$ time windows, with each window comprising $C + 1$ data points. The gradient descent algorithm is employed to optimize the model parameters until convergence is achieved.

3. For $t \in [N - C + 1, N + L]$, draw $M$ diffusion samples to compute the predicted mean $\mu_t$ and standard deviation $\sigma_t$ via Eq. (20) and Eq. (21). Estimate the prediction error variance $\sigma^2$ on the training set according to Eq. (23). If $|\Delta\sigma^2| \leq \tau$(with $\tau = 0.02$), proceed to step 6; otherwise continue to step 4.

4. The calculation of the outlier probability $P_o(x_t|\mu_t, \sigma_t^2, \sigma^2)$ for each data point is to

be conducted in accordance with Eq. (25). Aggregate these probabilities to obtain a dataset-level quality evaluation score using Eq. (26).

5. Data points exhibiting an outlier probability greater than 0.5 are identified as outliers. The corresponding data points in the training set are replaced with their predicted mean, and the imputation mask is updated, with the corresponding position elements updated to 1. Subsequent to the updating of the training set, the process is to revert to step 2. This process is repeated iteratively until $\Delta\sigma^2 \leq \tau$.

6. The final low-quality dataset, designated $\mathbf{U}$, and the clean time series dataset, marked $\mathbf{D}'$, are obtained after the data cleaning.

**Implementation details.** Additional variable, imputation mask $\mathbf{m}$, is introduced to flag whether the backfill value generated by the model prediction exists during the training process. If present, the loss for the corresponding training window is weighted by 0.5, mitigating self-confirmation bias.

## 3. Illustrative applications and discussion

In this section, two illustrative examples are presented to validate the effectiveness and practicality of the proposed method: one using suspension bridge monitoring data and the other using high-speed rail track monitoring data. These examples demonstrate the robustness of the model to outliers, its accuracy in data quality assessment, and its enhanced predictive performance after the identification and removal of outliers. It is imperative to acknowledge that these applications presuppose the system under investigation is time-invariant, signifying that subsequent to the model's training, its properties remain unchanged over time. Finally, we discuss potential issues that may arise when deploying the proposed method in real-world applications.

## 3.1 Illustrative application: bridge monitoring data
### 3.1.1 Dataset establishment

The data employed in this example are collected from a monitoring system installed on a suspension bridge in southwest China during the period from December 2023 to March 2024, as shown in Fig 5. The dataset primarily includes measurements of wind speed and wind direction. Three-dimensional ultrasonic anemometers, installed at the mid-span of the bridge, are used for data collection with a sampling frequency of 10 Hz. As illustrated in Fig. 6, the time-history of horizontal wind speed, horizontal wind direction, vertical wind speed, and vertical wind direction are presented in the four subplots, respectively. In the absence of extreme weather conditions, such as typhoons, wind direction data typically exhibits a dominant wind direction, accompanied by gentle and random changes related to the intensity of air turbulence (Song et al., 2017; Hou et al., 2019). It is evident from the plots that there are some significant mutation points in the horizontal wind direction data monitored at the millisecond level, which are divergent from the normal mutation caused by the measurement range from 360° to 0°. Consequently, effective data quality assessment for the monitoring data is a critical issue that warrants thorough investigation.

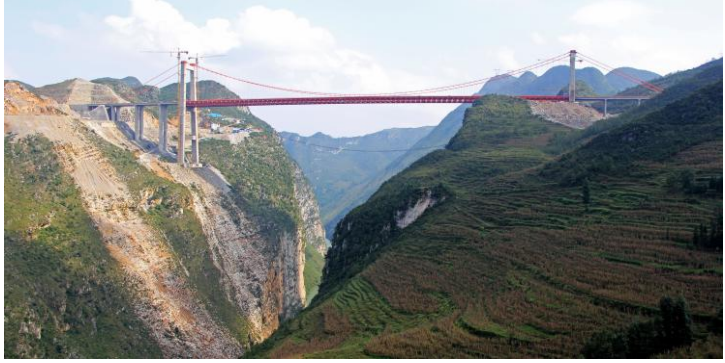
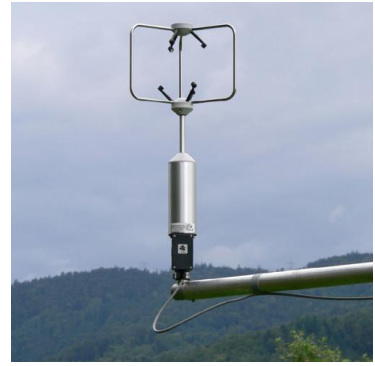

(a)                                (b)

**Fig. 5** Monitoring object and sensor:(a) suspension bridge (b) three-dimensional ultrasonic anemometer

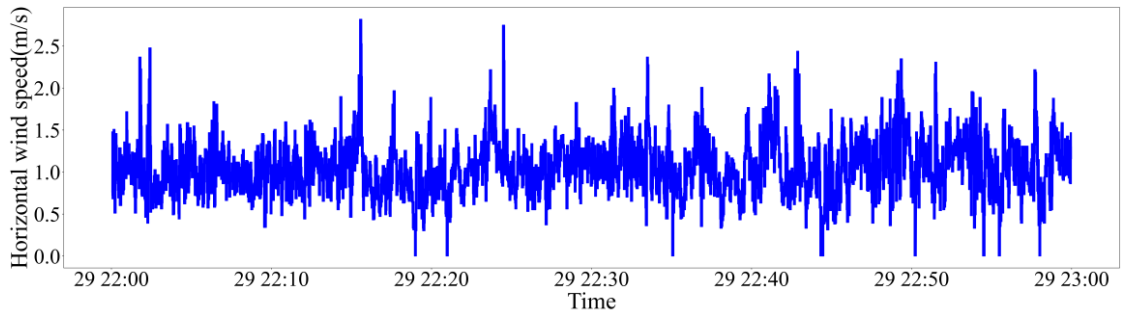

(a)

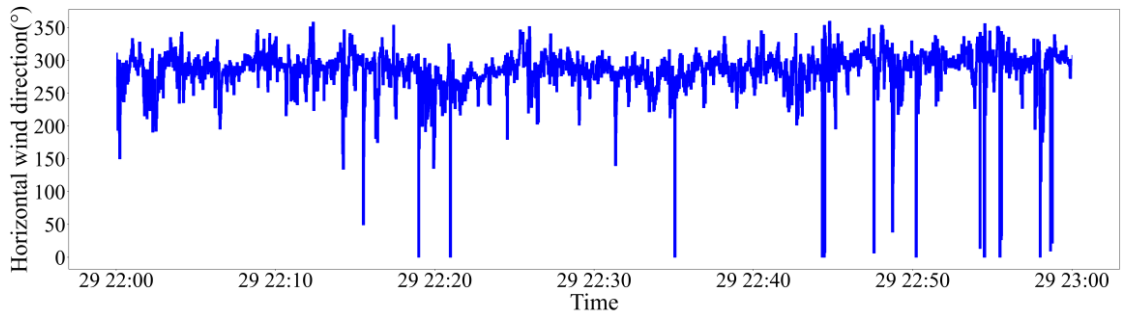

(b)

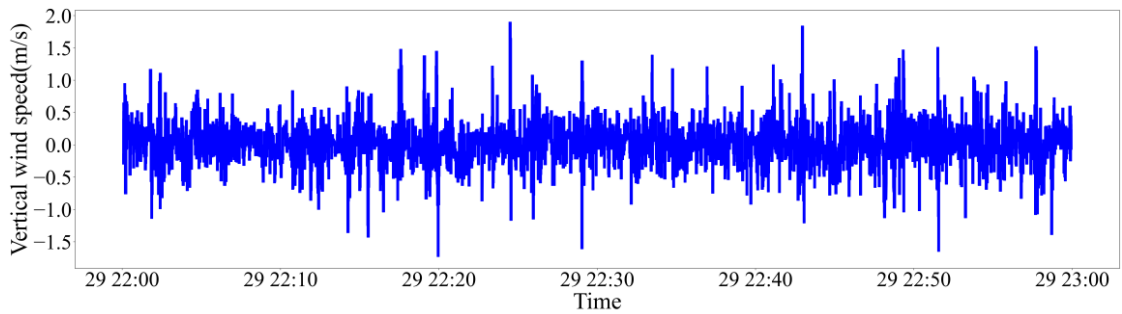

(c)

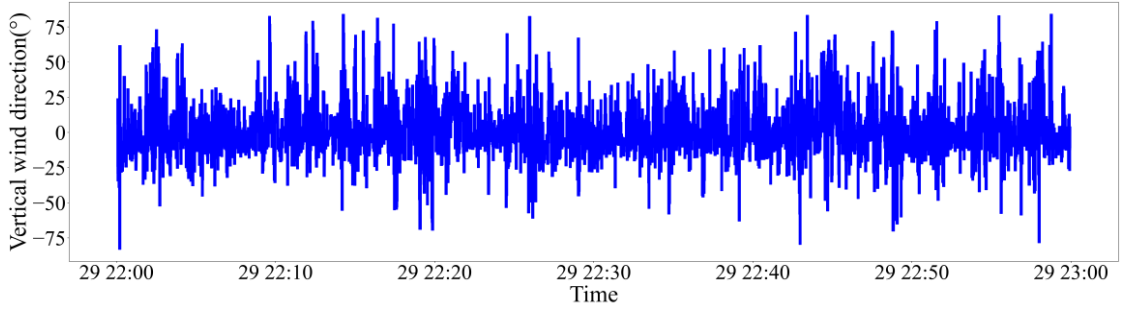

(d)

**Fig. 6** Wind speed and wind direction time-history curves: (a) horizontal wind speed, (b) horizontal wind direction, (c) vertical wind speed, (d) vertical wind direction

### 3.1.2 Model performance verification

In order to validate the performance of the proposed method in resisting outliers, wind direction monitoring over a one-hour period, comprising 28290 data points, was taken as input. And the data points have been labelled based on expert knowledge. The computations were performed on a computer device with an Intel i9-10900F CPU and an NVIDIA RTX 3070 GPU, utilizing PyTorch framework for implementation. The parameter settings for model training were as follows: the number of diffusion steps was set to $T = 140$, with maximum and minimum noise coefficients of $\beta_{max} = 0.1$ and $\beta_{min} = 0.0001$, respectively. The context window length was configured as $C = 80$, and the model underwent a total of 20 training epochs, using the Adam optimizer with an initial learning rate of 0.001. A cosine annealing strategy was employed for dynamically adjusting of the learning rate. The minimum learning rate was set to $10^{-9}$. With regard to data partitioning, 70% of the dataset was designated as the training set, while the remaining 30% was allocated to the test set. To evaluate the model's robustness against outliers, annotated outliers in the training set were incorporated during model training. However, outliers in the test set were excluded during evaluation to ensure the assessment focused on the model's prediction performance.

A series of ablation experiments were conducted using the following variations of the model: the original diffusion model (Base), the model with only the conditional embedding module (Conditional), the model with only the Huber loss function (Loss), the model with only the modified normalization method (Scaling), the model incorporating both normalization and loss function modifications (Scaling+L), and the model with both normalization and conditional embedding (Scaling+C). These experiments have been conducted in accordance with the parameters and dataset previously delineated.

In order to further quantify the predictive performance of the proposed model, two evaluation metrics, mean square error (MSE) and the logarithmic mean absolute error (LMAE), are computed as follows:

$$\text{MSE} = \frac{1}{N_t}\sum_{i=1}^{N_t}(y_i - \hat{y}_i)^2 \tag{27}$$

$$\text{LMAE} = \frac{1}{N_t}\sum_{i=1}^{N_t}|\log(y_i + 1) - \log(\hat{y}_i + 1)| \tag{28}$$

where $N_t$ is the total number of test data samples after the removal of outliers., and $y_i$ and $\hat{y}_i$ denote the actual and predicted wind direction values, respectively. The results are summarized in Fig. 7. It is evident that, among the three measures implemented to mitigate the impact of outliers, the incorporation of the conditional network module or the utilization of the Huber loss function on their own only lead to marginal improvements in prediction performance. In contrast, the quartile normalization method significantly enhances the model's accuracy. When all three measures are combined, the model achieves the best predictive performance, highlighting the importance of these strategies in improving the effectiveness of data quality assessment based on prediction deviation.

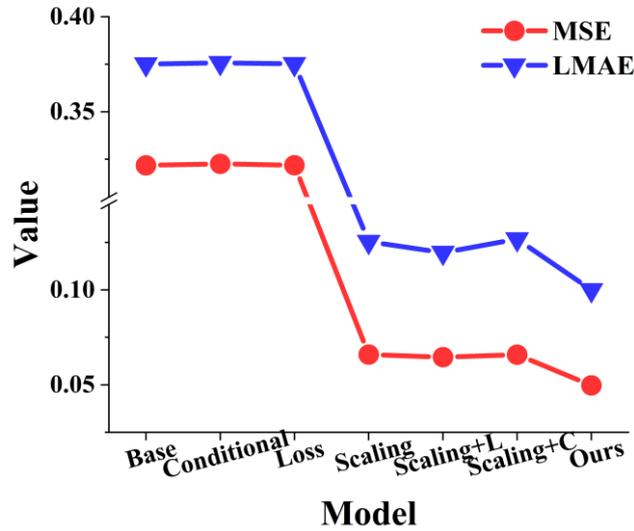

**Fig. 7** Model prediction performance with different models

In the subsequent analysis, we have investigated the effect of the diffusion step ($T$) and the context window length ($C$) on the predictive performance of the model. The number of diffusion steps directly affects the ability of the model to capture underlying data patterns. If $T$ is too small, the data does not undergo sufficient corruption into pure noise, limiting the model's ability to extract meaningful structure. Conversely, an excessively large $T$ leads to a prolonged sampling process during prediction, reducing computational efficiency. The results of varying $T$, shown in Fig. 8, indicate that performance metrics exhibit considerable variability with increasing $T$. The model achieves optimal performance at $T = 140$ and $T = 240$. Considering the trade-off between performance and efficiency, $T = 140$ is selected as the optimal choice. A similar analysis is conducted to assess the effect of context window length ($C$) on model performance, with results shown in Fig. 9. It can be found that increasing the context window length does not always improve prediction accuracy. The model achieves its best performance at $C = 80$.

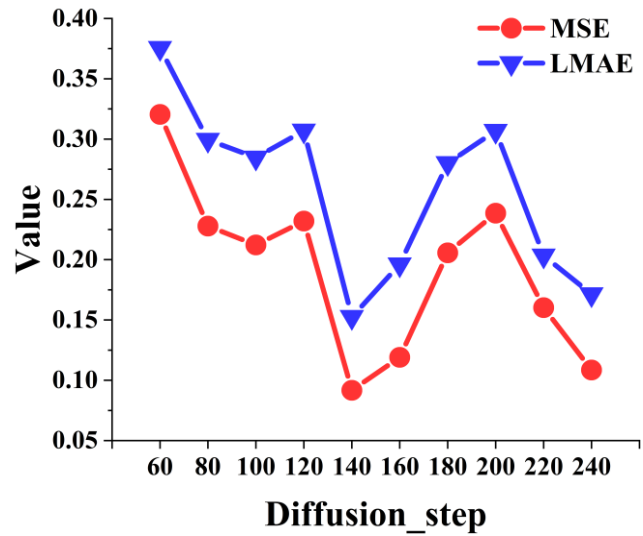

**Fig. 8** Model prediction performance with different number of diffusion step

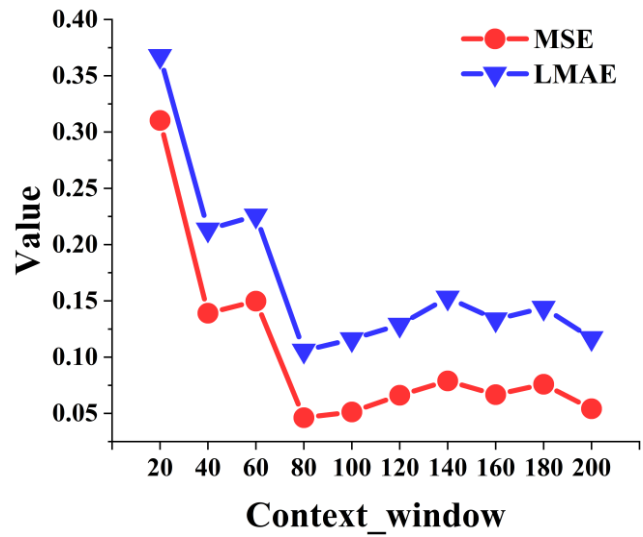

**Fig. 9** Model prediction performance with different number of context window

### 3.1.3 Data quality assessment performance verification

This subsubsection adopts the algorithm in Subsection 2.4 for data quality assessment. The initial training parameters are equivalent to those delineated in subsubsection 3.1.2. During the subsequent iterative update stage of model training, the model parameters are optimized based on the results of the previous iteration as the initial state. The initial learning rate in each iteration is reduced to 0.3 times that of the previous one, and the epoch is set to 10. The probabilistic data quality assessment results of the time series are displayed in Fig. 10. For the period in Fig. 10(a), a total of 611 outliers were detected by the proposed algorithm, accounting for 2.16% of the total data. The presence of low-data quality results in predictions that are significantly different from the actual values. As shown in Fig. 10, the greater the disparity between the predicted and actual values, the darker the data point appears, indicating a higher degree of "outlier-ness" and a larger calculated outlier probability.

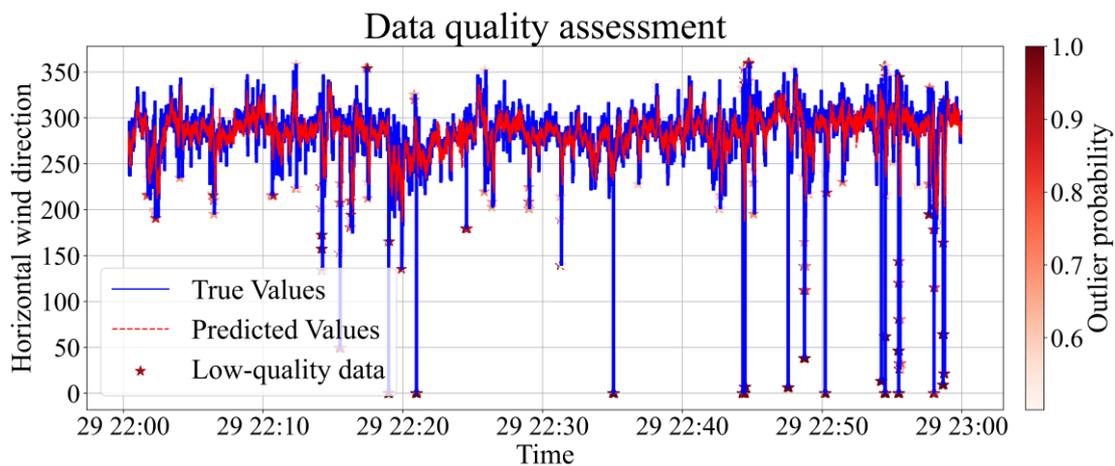

(a)

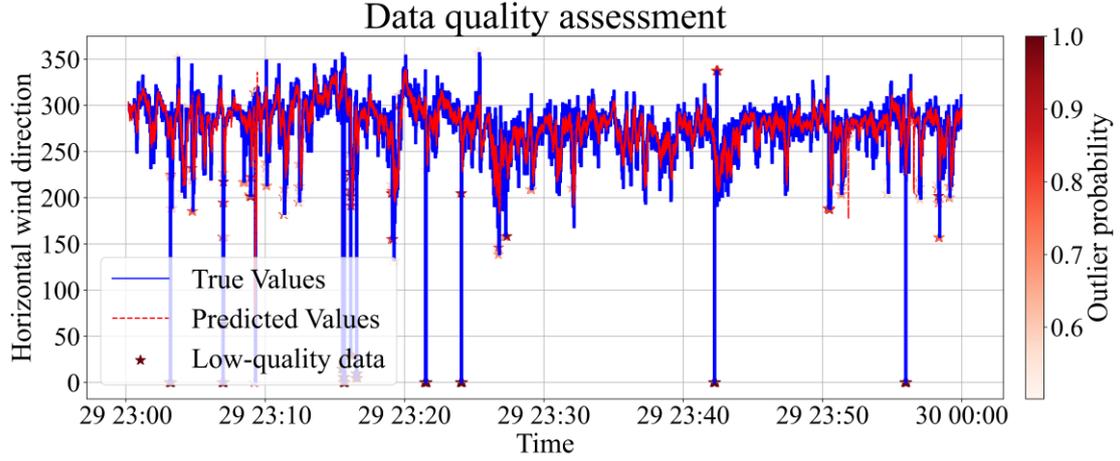

(b)

**Fig. 10** Probabilistic data quality assessment result of the representative time series

Generally, in meaningful structural health monitoring data, the proportion of outlier is relatively small. The selection of appropriate metrics to quantify outlier detection performance is of paramount importance. This paper employs three metrics, namely, Precision, Recall and F1-score as follows:

$$\text{Precision} = \frac{TP}{TP+FP} \tag{29}$$

$$\text{Recall} = \frac{TP}{TP+FN} \tag{30}$$

$$F1-\text{score} = \frac{2\times \text{Precision}\times \text{Recall}}{\text{Precision}+\text{Recall}} \tag{31}$$

where TP, FP and FN denote the numbers of true-positive, false-positive and false-negative samples, respectively. A higher Precision is indicative of a lower proportion of normal data misidentified as outlier (a phenomenon referred to by some scholars as "swamping") (Yuen & Mu, 2012), thus implying a lower likelihood of misclassification. Conversely, higher Recall is indicative of a higher proportion of true outlier being correctly identified by the method (a phenomenon referred to as "masking") (Yuen & Mu, 2012), thereby implying a lower likelihood of missing detections. The F1-score is defined as the harmonic mean of Precision and Recall, serving as a comprehensive metric that takes both factors into account.

We have compared our method with several strong baselines that are representative of data quality assessment and anomaly detection, including DBSCAN (Aslan & Onut, 2022), Local Outlier Factor (LOF) (Tan et al., 2024), Isolation Forest (Iso) (Liu et al., 2008), Anomaly Transformer (ATF) (Xu et al., 2021). Furthermore, a comparison of the present method with the percentile-based predictive error variance parameter estimation approach (Percentiles) (Yuen & Ortiz, 2017) was undertaken. It is important to note that DBSCAN, LOF and Iso are machine learning methods, whereas ATF is a deep learning method, implying that this comparison quantitatively reveals the superior performance of the proposed method compared to the baseline machine and deep learning methods. All experiments were conducted under three settings to ensure a fair comparison. First, DBSCAN, LOF and Iso were applied to the prediction residuals produced by the conditional diffusion model, rather than to the raw monitoring data. This choice is motivated by two fairness considerations. Our method performs probabilistic scoring in the residual domain, and applying baselines directly to raw measurements would confound temporal dynamics modeling with outlier detection. Evaluating all methods on the same residual sequence ensures that they are assessed using an identical anomaly-evidence signal. In addition, raw SHM time series typically exhibit strong autocorrelation, which may substantially degrade density-based and isolation-based detectors when used without an explicit temporal model. Using the CDM-generated residuals for all methods therefore isolates the outlier-detection capability itself, rather than disadvantaging baselines that do not include a dedicated forecasting component. Second, ATF similarly includes outliers during the model training process. Third, the hyperparameters used in these methods were determined through grid search to ensure the best performance of each technique.

The results of the three metrics in Eq. (29)-(31) are demonstrated in Fig. 11. As can be seen from the figure, the Precision and Recall metrics of the DBSCAN and the method using percentiles to estimate the predictive error variance parameter for data quality assessment are similar, indicating balanced model performance. LOF and Iso exhibit comparable performance, achieving up to 90% recall, but with precision around 50%. The ATF deep learning method performs exceptionally well in terms of precision, but it has the lowest recall. This could be attributed to the relatively poor resistance of unsupervised data quality assessment methods based on deep learning models to outliers. These models typically use only normal data during the training phase, which further limits the practicality in engineering applications. In contrast, the method proposed in this paper achieves the best performance, with an F1 score of 85%, outperforming the other data quality assessment techniques.

To further assess computational efficiency, we compared the runtime required for a single round of anomaly diagnosis across these methods. All runtimes were measured using Python's Time module, and each method was executed five times with the average runtime reported; the results are summarized in Table. 1. The ATF method was excluded because it performs anomaly diagnosis after retraining a deep network, resulting in a runtime that is substantially (several orders of magnitude) higher than that of the other methods. As shown in Table.1, although our approach is the most time-consuming among the remaining three methods, it requires only 3.41s on average to compute the anomaly probabilities for 28130 data points, which remains acceptable in practice.

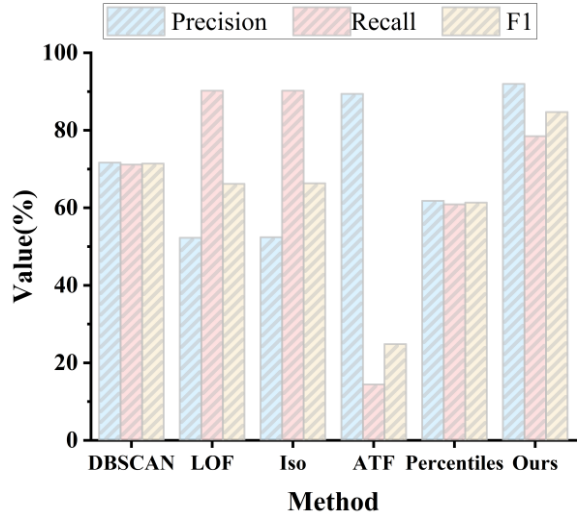

**Fig. 11** Performance metrics with different data quality assessment methods

**Table. 1** Runtime of outlier diagnosis using different methods

| Method | Runtime(s) |
| --- | --- |
| DBSCAN | 0.61 |
| LOF | 3.07 |
| Iso | 1.02 |
| Ours | 3.41 |

The changes in the *QES* indicators for the two time-series datasets corresponding to Fig. 10 after each round of iterative cleaning are presented in Fig. 12. As outlier identification progresses, *QES* increases consistently and eventually reaches an optimal level. Different values of $k$ lead to only a moderate vertical shift in *QES* during the early iterations, while the overall behavior remains unchanged: *QES* increases monotonically as the iterative cleaning proceeds, and the curves converge as the algorithm stabilizes. This indicates that the qualitative conclusions drawn from *QES* are insensitive to the choice of $k$, supporting the validity and rational of the proposed indicator definition. Moreover, the proposed algorithm typically converges within a small number of iterations, generally no more than ten.

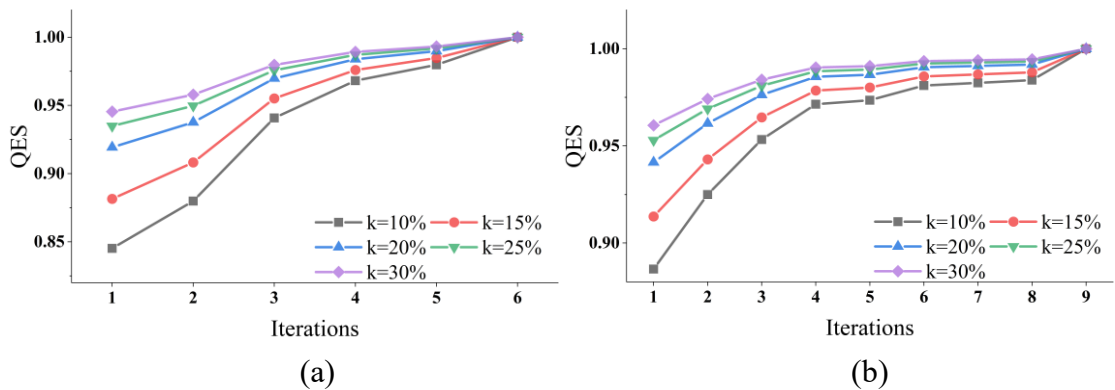

**Fig. 12** Evolution of *QES* over iterations under different values of $k$

## 3.2 Illustrative application: high-speed railway track monitoring data

In this example, monitoring data of high-speed railway track structures from the period between August 2018 and December 2018 is utilized. The types of monitoring data include track temperature and strain collected using fiber Bragg grating sensors. This monitoring system mainly monitors the impact of temperature effects on track. To avoid the influence of train operation, the sampling frequency is 10 min. More relevant details can be obtained from Li et al. (2024). The number of data samples is about 17577 points. The track temperature and strain time-history curves of some example measuring points are shown in Fig. 13. The track strain data should exhibit a periodic characteristic in response to changes in temperature. However, as is evident in the figure that there are some abrupt points (outliers) in the strain monitoring data.

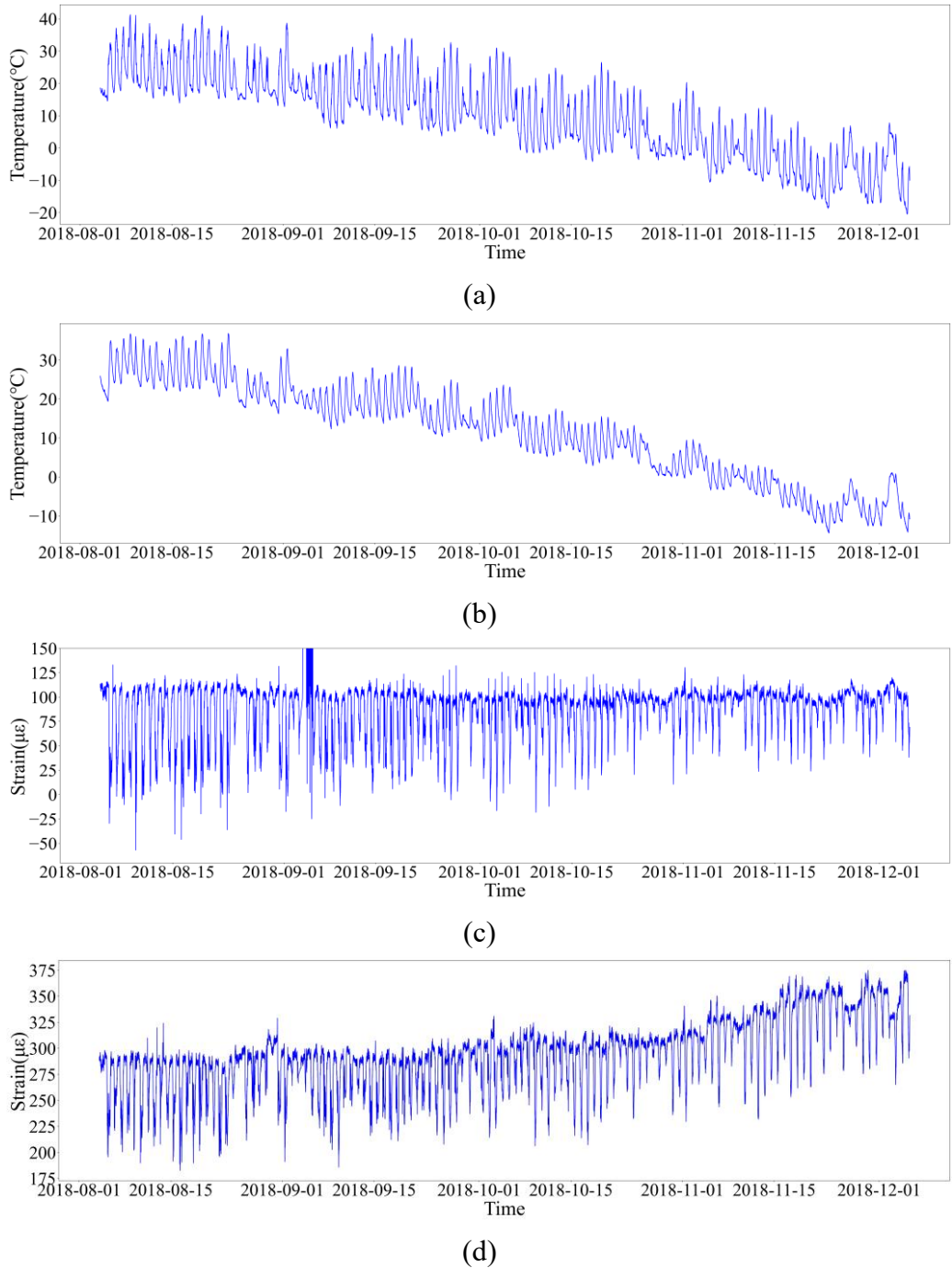

**Fig. 13** Time-history curves: (a)(b) track temperature, (c)(d) track strain

The results of the data quality assessment on two exemplary time series are presented in Fig. 14, where the parameter settings are the same as that in subsubsection 3.1.3. A total of 507 outliers was detected by the proposed algorithm, accounting for 2.88% of the total data. The true values are indicated by the solid blue line, while the model's predicted values are shown by the dashed red line. The outliers have been marked by the pentagram, where the darker the star, the higher the outlier probability for that specific point. To provide a more detailed illustration, the data contained within the red rectangular box in the figure has been magnified for closer inspection. As demonstrated in the figure, the greater the deviation between the predicted values and the true values, the darker the color of the identified outliers. This suggests that the definition of outlier probability is valid, as it effectively quantifies the degree of anomaly of the data points.

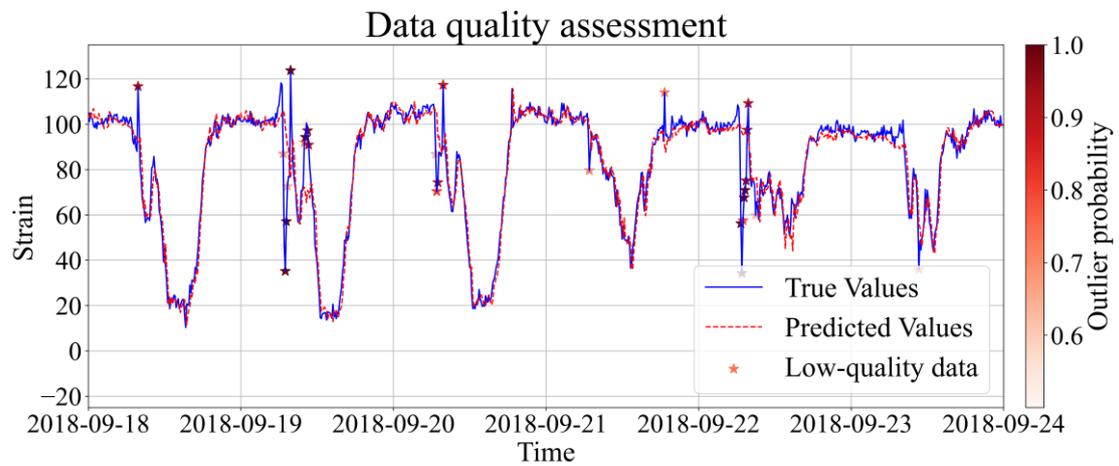

(a)

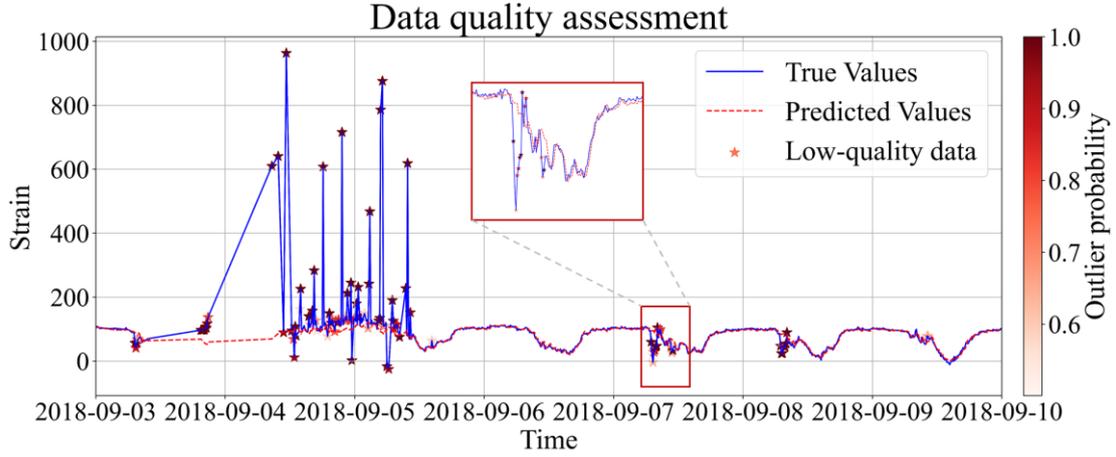

(b)

**Fig. 14** Probabilistic data quality assessment result of the representative time series

**Table. 2** Metrics before and after outlier detection and removal

| Time series | Metric | Before outlier removal | After outlier removal |
| --- | --- | --- | --- |
| 1 | MSE | 0.0030 | 0.0024 |
| 1 | LMAE | 0.0210 | 0.0202 |
| 2 | MSE | 0.0042 | 0.0011 |
| 2 | LMAE | 0.0112 | 0.0056 |

The comparison of the prediction results before and after data quality assessment and removal in two different time series has been conducted. The difference is that identified outliers are removed when prediction is performed after data quality assessment. In Table. 2, the corresponding MSE and LMAE values for the data before and after data quality assessment is given. The values of both indicators become smaller after the outlier identification and removal. It is interesting to see that for time series 2, the prediction indicators significantly decreased. This suggests that our algorithm can successfully identify outliers even when there are serious data anomalies.

At the final stage of this study, we examined how the *QES* metrics for the two representative time series data corresponding to Fig. 14 changed after each cleaning iteration. The results are presented in Fig. 15. We can draw a similar conclusion that the *QES* metric typically reaches an optimum within fewer than ten iterations, indicating that all possible outliers have been identified. And the qualitative conclusions drawn from *QES* are insensitive to the choice of $k$. These results further validate the effectiveness of the proposed method.

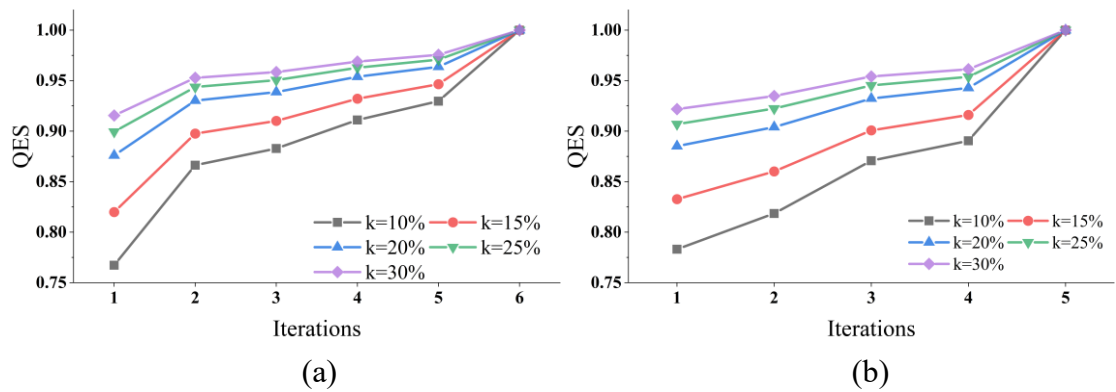

**Fig. 15** Evolution of *QES* over iterations under different values of $k$

## 3.3 Discussion

In this study, a novel method for SHM data quality assessment is developed. Notwithstanding its effectiveness in outlier diagnosis and data cleaning, the proposed framework raises several practical considerations that deserve further discussion.

First, the illustrative case studies adopt a time-invariance assumption after model training to demonstrate the effectiveness of the proposed data quality assessment approach. However, real-world structures are subject to continuous aging, which introduces shifts in the underlying distribution of the data. These shifts, often referred to as concept drift, can be gradual and persistent rather than abrupt, as seen in transient outliers caused by sensor faults or environmental disturbances. Under the time-invariant assumption, natural degradation may indeed be erroneously identified as outliers. To address this challenge, we recommend adopting a trigger-based model update strategy in practical deployments. Since natural aging typically manifests as a slow, persistent distributional shift rather than isolated spike-like anomalies, frequent retraining is unnecessary. A more practical approach is to treat phenomena such as pronounced temporal clustering of detected anomalies as a potential signal of "model-system mismatch", and then trigger an update of the model using recently collected, quality-screened data. In this way, the model can gradually track the evolving structural state and avoid misinterpreting long-term, gradual aging as a data-quality issue.

Second, we evaluate the computational efficiency of the proposed algorithm and its feasibility for real-time deployment. In the proposed framework, the dominant computational cost arises from pointwise probabilistic forecasting, namely estimating the predictive distribution through reverse diffusion sampling, rather than from model training or the subsequent computation of anomaly probabilities. In our current Python implementation, timed using the Time module, a complete training run under the aforementioned settings takes approximately 600s, and evaluating outlier probabilities for all predicted data points requires 3.41s. In contrast, estimating pointwise predictive distributions for 28130 data points with $T = 140$ diffusion steps and 100 samples requires 3754.41s. This corresponds to an average throughput of approximately 7.5 data points per second, or about 0.13s per data point. This runtime is sufficient for many SHM scenarios with low-to-moderate sampling rates, such as minute-level or 10-mintue sampling, where near-real-time alerts can be generated reliably. For high-frequency data streams, however, a straightforward deployment strategy that performs full sampling for every data point may become computationally demanding, and practical acceleration strategies are therefore required. Overall, these timing results demonstrate that the proposed method is practical for offline analysis and for many online SHM settings. Real-time deployment under high-frequency sampling is expected to be achievable by incorporating standard inference-speed optimizations, which we identify as an important direction for future work.

Finally, we discuss the rationale for applying the 0.5 threshold to datasets with different noise-to-signal ratios. In our framework, outlier classification is conducted in the probability domain rather than by applying a hard residual cutoff. Specifically, a data point is flagged as an outlier when its outlier probability $P_o$ exceeds 0.5. The value 0.5 corresponds to the standard Bayes/MAP decision rule under the common assumption that false-positive and false-negative costs are approximately symmetric. More importantly, $P_o$ is computed from a predictive distribution that explicitly incorporates both the model predictive uncertainty $\sigma_t^2$ and a resampling-based estimate of the prediction-error variance $\sigma^2$, thereby reflecting different noise levels and modeling errors in a data-driven manner. Consequently, when the noise-to-signal ratio changes, the probability-based criterion will be adjusted accordingly, therefore, the same threshold does not act as a strict, global "residual cutoff" across the entire dataset. This robustness is also supported empirically by our case studies. We validated the method on real monitoring data from a suspension bridge and a high-speed railway track, which differ substantially in structural form, operating environment, sensor type, and sampling characteristics, and thus typically exhibit different noise-to-signal ratios. Despite these differences, the proposed probability-based criterion with the default threshold $P_o > 0.5$ remained effective in both applications, suggesting that the threshold is reasonably transferable across heterogeneous SHM datasets. That said, if a particular deployment involves asymmetric decision costs (e.g., prioritizing extremely low false alarms or maximizing sensitivity), the threshold should be adjusted upward or downward as appropriate to meet the application requirements.

## 4. Conclusion

This study proposes a prediction deviation-based method for assessing the quality of structural health monitoring data using a univariate implicit autoregressive model, thereby enabling outlier diagnosis and data cleaning. By incorporating context through conditional embedding, applying quartile normalization, and using a Huber loss function, the proposed framework effectively learns the underlying data distribution, even in the presence of outliers. Additionally, this framework can be considered an implicit autoregressive prediction model. The outlier probability for each data point is computed within this framework, and these probabilities are then aggregated into a global quality evaluation score for dataset-level assessment.

As a univariate implicit autoregressive method based on prediction deviation, it offers the following distinct advantages. First, this framework is inherently outlier-resistant, removing the need for pre-cleaning or noise-free training data, and is therefore highly adaptable to complex and uncertain real-world engineering environments. Second, unlike existing methods that only categorize data as normal or anomalous, our framework quantifies the degree of anomaly for each data point, providing a comprehensive data-quality assessment at both the local (pointwise outlier probability) and global ($QES$) levels. Third, the method demonstrates broad applicability, consistently identifying outliers across various structures and heterogeneous sensor types. These advantages suggest that the proposed method is well-suited for integration into operational SHM systems.

For future study, it would be useful to relax the assumption of a time-invariant system and extend the present univariate framework to non-stationary and multivariate SHM data. Further gains may come from improving computational efficiency, for example by optimizing the diffusion sampling process, to enable real-time deployment in large-scale SHM systems. Despite these limitations, the proposed method constitutes a significant advance in SHM data management, providing a robust probabilistic tool for quantifying and enhancing data quality and thereby improving the reliability of SHM outcomes.